\documentclass{ieeeaccess}
\usepackage{cite}
\usepackage{amsmath,amssymb,amsfonts}
\usepackage{algorithmic}
\usepackage{graphicx}
\usepackage{textcomp}
\usepackage{subcaption} 
\usepackage{color} 

\def\BibTeX{{\rm B\kern-.05em{\sc i\kern-.025em b}\kern-.08em
    T\kern-.1667em\lower.7ex\hbox{E}\kern-.125emX}}
\begin{document}
\history{Date of publication xxxx 00, 0000, date of current version xxxx 00, 0000.}
\doi{10.1109/ACCESS.2017.DOI}

\title{An Environment-Adaptive Position/Force Control Based on Physical Property Estimation}
\author{\uppercase{Tomoya Kitamura}\authorrefmark{1, 2}, \IEEEmembership{Member, IEEE},
\uppercase{Yuki Saito}\authorrefmark{2}, \IEEEmembership{Member, IEEE},
\uppercase{Hiroshi Asai}\authorrefmark{2}, \IEEEmembership{Member, IEEE},
\uppercase{Kouhei Ohnishi}\authorrefmark{2}, \IEEEmembership{Life Fellow, IEEE}.}
\address[1]{Department of Electrical Engineering Faculty of Science and Technology, Tokyo University of Science, Noda, Japan}
\address[2]{Haptics Research Center, Keio University, Kanagawa, Japan 
(e-mail: t.kitamura@rs.tus.ac.jp, ysaito@haptics-c.keio.ac.jp, h-asai@haptics-c.keio.ac.jp, ohnishi@sd.keio.ac.jp)}

\tfootnote{The part of this work has been supported by the JST-Mirai Program Grant Number JPMJMI21B1, Japan.}

\markboth
{T. Kitamura \headeretal: An Environment-Adaptive Position/Force Control Based on Physical Property Estimation}
{T. Kitamura \headeretal: An Environment-Adaptive Position/Force Control Based on Physical Property Estimation}

\corresp{Corresponding author: Tomoya Kitamura (e-mail: t.kitamura@rs.tus.ac.jp).}
.

\begin{abstract}
The current methods to generate robot actions for automation in significantly different environments have limitations. 
This paper proposes a new method that matches the impedance of two prerecorded action data with the current environmental impedance to generate highly adaptable actions. 
This method recalculates the command values for the position and force based on the current impedance to improve reproducibility in different environments. 
Experiments conducted under conditions of extreme action impedance, such as position and force control, confirmed the superiority of the proposed method over existing motion reproduction system. 
The advantages of this method include the use of only two sets of motion data, significantly reducing the burden of data acquisition compared with machine-learning-based methods, and eliminating concerns about stability by using existing stable control systems. 
This study contributes to improving the environmental adaptability of robots while simplifying the action generation method.

\end{abstract}


\titlepgskip=-15pt

\maketitle

\section{Introduction}
Recently, the importance of robotic technology has been increasingly acknowledged across various sectors, including industry, domestic applications, and healthcare\cite{c22, c23, c24}. 
For robots to operate effectively in these diverse settings, they must adapt to objects with different shapes and stiffness levels. 
This often involves understanding the object's impedance, which combines the force and position, making impedance adjustment a key challenge in robot control. 
Control operations must address both position and force dimensions to adapt seamlessly to the surrounding environment.

Various hybrid control strategies incorporating acceleration have been proposed to manage both position and force dimensions simultaneously \cite{c1, c2}. These strategies utilize the acceleration reference values generated by two distinct controllers: one dedicated to the position and the other to the force. Furthermore, motion reproduction systems (MRS) have been developed to steer robotic actions using historical position and force data to accurately replicate these actions\cite{c3, c4}. However, a significant issue arises when the recorded motion differs substantially from the object's environment during reproduction, causing the motion to become unstable. Such environmental changes can be classified into two main categories: differences in the object's position and physical properties, both of which require distinct approaches for handling. A method utilizing image information to correct the motion has been proposed for cases where the object's position differs\cite{c40, c41, c42}. Conversely, this issue has not been sufficiently addressed when the object's physical properties differ from those at the time of recording. In response, some studies suggest that differences in physical properties can be managed by adjusting control stiffness\cite{c43, c44}, while others propose sequentially estimating the object's impedance and modifying the motion accordingly\cite{c46, c47, c45}. However, these methods face the challenge of determining whether the recorded position or force should be prioritized during reproduction. Typically, one command value is fixed while the other is adjusted to accommodate fluctuations in physical properties. Although this improves adaptive performance, it does not ensure faithful reproduction of the recorded motion. Therefore, it is essential to accurately estimate the motion to be reproduced and generate appropriate position and force command values tailored to the object's physical properties, which may differ from those at the recording time. This approach enables more accurate reproduction of the recorded motion while maintaining robust adaptation to environmental changes.

There is ongoing research into approaches using machine learning for environmental adaptation. Reinforcement learning is a method for generating robot movements to execute tasks, where the learning device's output typically includes the robot's position or joint stiffness, contributing to improved environmental adaptability\cite{c5, c6, c7, c8}. However, reinforcement learning requires a significant number of learning movements, leading to high temporal and equipment costs. Alternatively, imitation learning has given robots human-like adaptability by learning human movements\cite{c9, c10, c11, c12}. This approach is more efficient, requiring less learning data than reinforcement learning, but still necessitates multiple sets of motion data. It has been reported that the success rate of tasks is high for all machine learning-based methods and that environmental adaptability has also improved\cite{c5, c6, c7, c8, c9, c10, c11, c12}. However, machine learning methods generally involve long motion generation times due to large sampling intervals. Additionally, these methods require careful adjustment of the learning device's hyperparameters based on the task and robot. In contrast, MRS leverages pre-recorded motions, enabling faster motion generation. Therefore, if motions for MRS that adapt to the environment can be generated using a small amount of motion data, it will be possible to achieve quick and efficient environmental adaptation.

In this study, we propose a new approach for matching the impedance of the environment and motion. 
In this method, we recorded motion data from a motor interacting with two samples with different environmental impedances. 
Overall, this study aims to improve the adaptability of robot motion under various environmental conditions using the proposed impedance matching method. 
The originality of this study is that, unlike machine learning methods, it uses only two data points, and unlike conventional MRS, it can adapt to different environments without losing stability.

\begin{table}[h!]
\centering
\caption{Nomenclature table for variables used in the study.}
\label{tb:Nomenclature} 
\begin{tabular}{|l|l|l|}
\hline
\textbf{Symbol} & \textbf{Description} & \textbf{Unit} \\ \hline
$x$ & Position & rad \\ \hline
$\dot{x}$ & Velocity & rad/s \\ \hline
$\ddot{x}$ & Acceleration & rad/s$^2$ \\ \hline
$f$ & Force & Nm \\ \hline
$\circ^{\mathrm{ref}}$ & Reference value & - \\ \hline
$\circ_{\mathrm{cmd}}$ & Command value & - \\ \hline
$\circ_{\mathrm{res}}$ & Response value & - \\ \hline
$\circ_{\mathrm{rec}}$ & Recorded value & - \\ \hline
$\circ_{\mathrm{p}}$ & Position controller-related value & - \\ \hline
$\circ_{\mathrm{f}}$ & Force controller-related value & - \\ \hline
$\circ_{\mathrm{A}}, \circ_{\mathrm{B}}, \circ_{\mathrm{C}}$ & Data for Sample A, B, C & - \\ \hline
$K_{\mathrm{pos}}$ & Proportional gain & 1/s$^2$ \\ \hline
$K_{\mathrm{vel}}$ & Differential gain & 1/s \\ \hline
$K_{\mathrm{for}}$ & Force feedback gain & - \\ \hline
$J_{\mathrm{n}}$ & Nominal inertia of motor & kg·m$^2$ \\ \hline
$f_{\mathrm{dis}}$ & Estimated disturbance force & Nm \\ \hline
$f_{\mathrm{rfob}}$ & Estimated reaction force & Nm \\ \hline
$\alpha_{\mathrm{A}}, \alpha_{\mathrm{B}}$ & Weights for motion data interpolation & - \\ \hline
$M$ & Mass & kg·m$^2$ \\ \hline
$D$ & Damper & Nm s/rad \\ \hline
$K$ & Spring & Nm/rad \\ \hline
$H$ & Load & Nm \\ \hline
$f^{\mathrm{noise}}$ &High-frequency noise signal & Nm \\ \hline
$f^{\mathrm{noise}}_{\mathrm{amp}}$ & Amplitude of noise signal & Nm \\ \hline
$den$ & Denominator used  & - \\
 & in interpolation calculations & \\ \hline
\end{tabular}
\end{table}

Six samples were prepared in each experiment. 
For two samples, data were recorded during the execution of either position or force control. 
The proposed method was then applied to the remaining four samples. 
A single-degree-of-freedom rotary motor was used in these experiments. 
The effectiveness of the proposed method was assessed under extreme conditions such as force control (i.e., zero-motion impedance) and position control (i.e., infinite motion impedance). 
Additionally, we compared the control performance of our method with that of the MRS. 
The results confirmed that our proposed method significantly enhanced the control performance in the position and force dimensions.
\begin{figure*}[t]
\centering
\includegraphics[width=110mm]{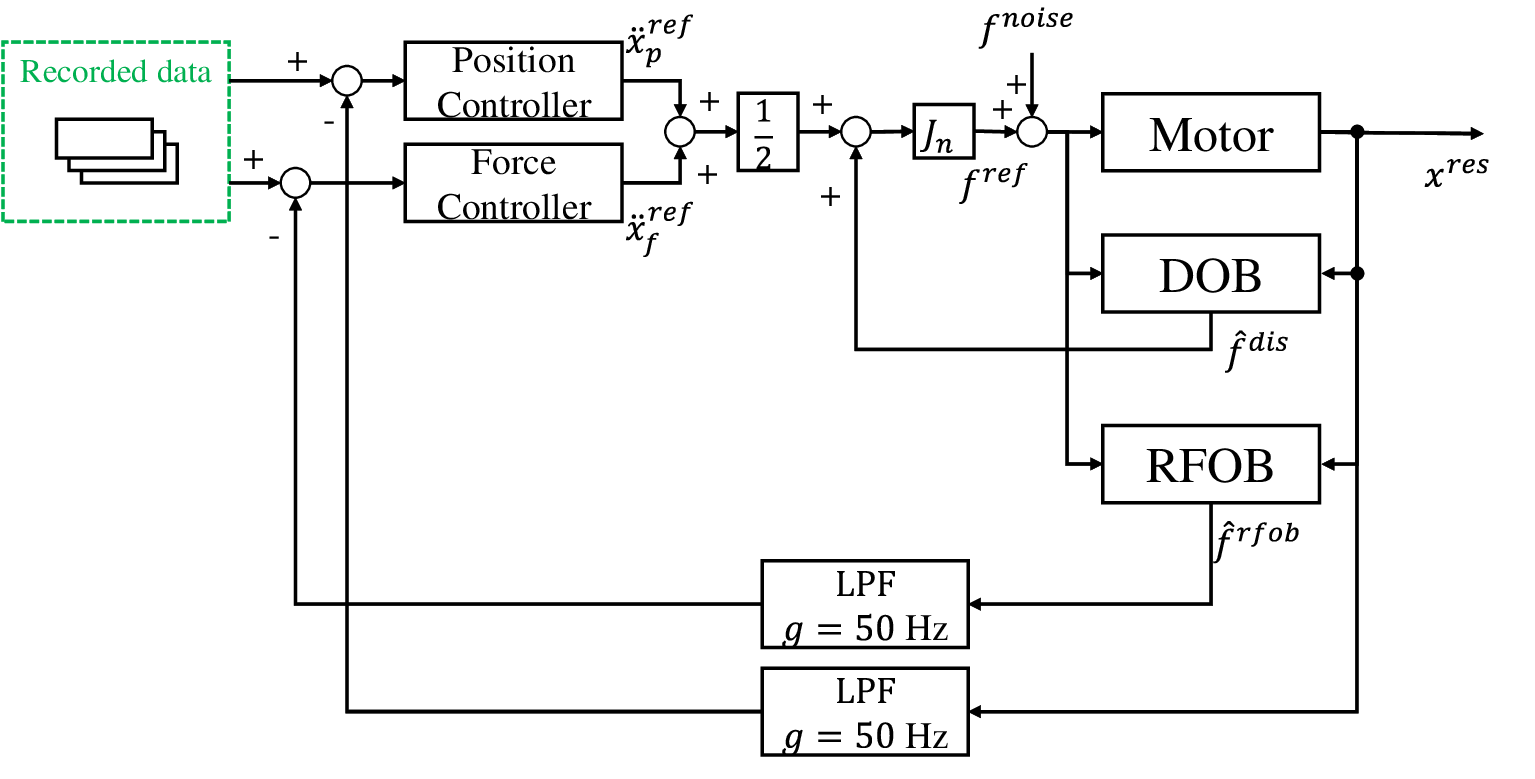}
    \caption{Block diagram with MRS}
    \label{fig:MRS} 
\end{figure*}

The primary advantage of this method lies in its ability to manage tasks of both position and force control within the same system while limiting the need for data acquisition to just two sets of motion data. This study makes the following contributions: 
\begin{itemize}
\item The proposed method achieved an approximately 70\% reduction in position errors for position control tasks and force errors for force control tasks compared to the traditional MRS. Generating command values and utilizing existing stable controllers ensured system stability without the need for additional stability analysis.
\item By estimating the motion impedance from two sets of data and aligning it with the environmental impedance, this approach reduces the amount of preliminary motion data required compared to machine-learning-based methods and simplifies the learning process. In addition, because it is applied to environments corresponding to the extrapolation of two actions, it has a broader application scope than machine learning methods.
\item This method characterizes motion skills using impedance parameters. Therefore, an abstract representation of behavior opens up potential applications in teaching skills to beginners, among other possibilities.
\end{itemize}

The remainder of this paper is organized as follows: Section II focuses on the methodology and provides detailed explanations of the hybrid control, MRS, physical property estimation, and the proposed methods. 
Section III introduces the experimental methodology and describes the experimental design and procedures for a single-degree-of-freedom (DOF) rotary motor. 
Section IV presents the experimental results and compares the effectiveness of the MRS and the proposed method. 
This includes a statistical verification using a t-test, providing a deep consideration of the utility of the proposed method. 
Finally, Section V concludes the paper, summarizes the key points of this study, and discusses future research directions. 
Table~\ref{tb:Nomenclature} presents the details of each variable, subscript, and abbreviation.

\begin{figure*}[t]
\centering
\includegraphics[width=120mm]{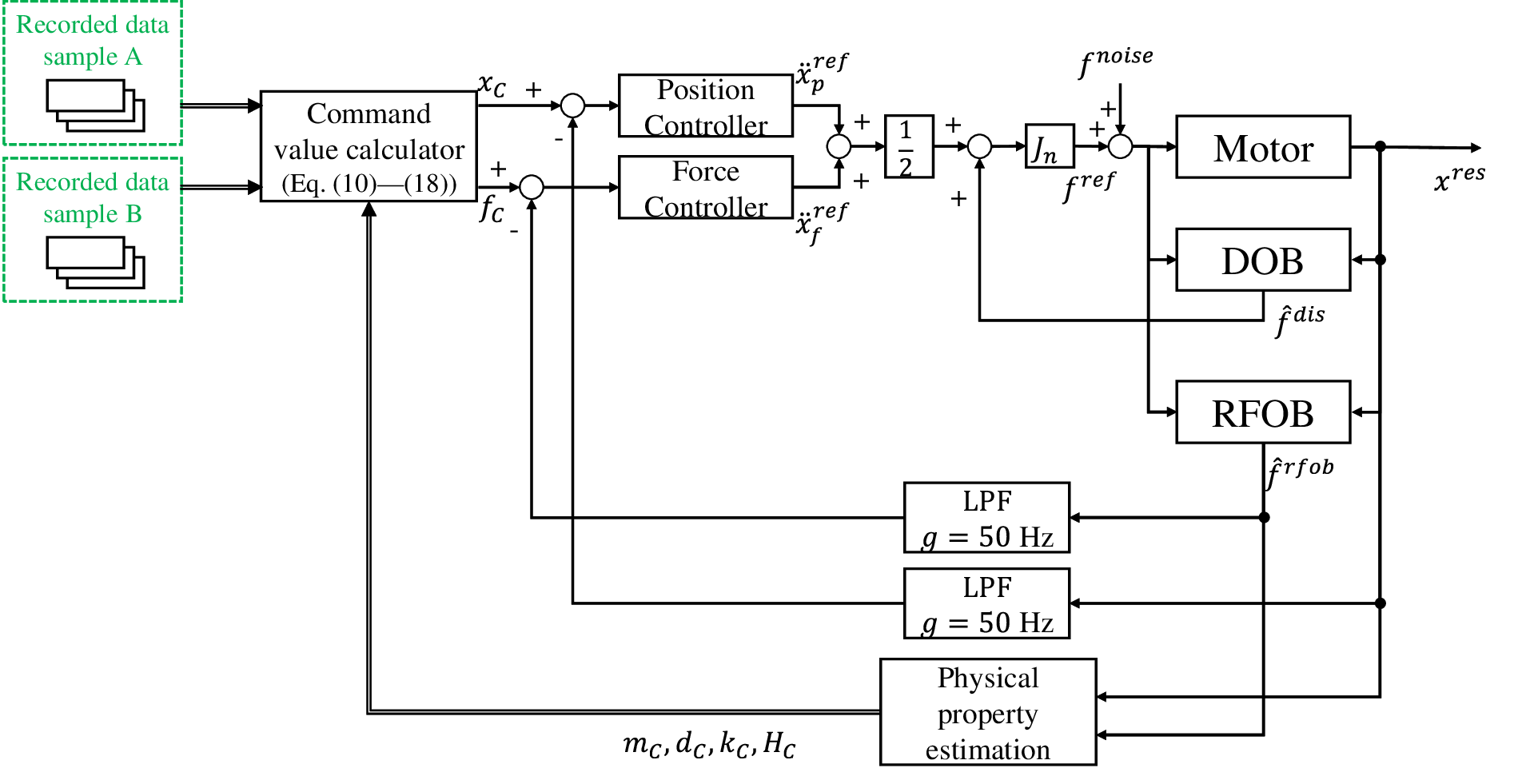}
    \caption{Block diagram with proposed method}
    \label{fig:Prop} 
\end{figure*}

\section{Methods}
\subsection{Control methods}
This section discusses position/force control, which orchestrates position and force control by utilizing acceleration reference values. It begins by elucidating the concepts of position and force control and then describes how position/force control combines these two aspects synergistically.
\subsubsection{Position control}
Position control is performed so the current position response value $x^{\mathrm{res}}$ approaches the desired position command value $x^{\mathrm{cmd}}$. Therefore, the acceleration reference value in position control $\ddot{x}^{\mathrm{ref}}_{\mathrm{p}}$ is described by (\ref{eq:eq1}).
\begin{align}
\ddot{x}^{\mathrm{ref}}_{\mathrm{p}}=(K_{\mathrm{pos}}+K_{\mathrm{vel}}\frac{d}{dt})(x^{\mathrm{cmd}}-x^{\mathrm{res}}), 
\label{eq:eq1}
\end{align}
where, $K_{\mathrm{pos}}$ and $K_{\mathrm{vel}}$ represent proportional and differential gains, respectively. Furthermore, by adding the estimated disturbance $\hat{f}^{\mathrm{dis}}$ estimated by the disturbance observer\cite{c21}, robust control against unknown disturbances is achieved.
\begin{align}
\ddot{x}^{\mathrm{ref}}_{\mathrm{p}}= (K_{\mathrm{pos}}+K_{\mathrm{vel}}\frac{d}{dt})(x^{\mathrm{cmd}}-x^{\mathrm{res}})+\hat{f}^{\mathrm{dis}}/J_{\mathrm{n}},
\label{eq:eq2}
\end{align}
where, $J_{\mathrm{n}}$ represents nominal inertia of motor.
\subsubsection{Force control}
The acceleration reference value in force control $\ddot{x}^{\mathrm{ref}}_{\mathrm{f}}$ to satisfy the desired force command value $f^{\mathrm{cmd}}$ is expressed by (\ref{eq:eq3}).
\begin{align}
\ddot{x}^{\mathrm{ref}}_{\mathrm{f}}=K_{\mathrm{for}}(f^{\mathrm{cmd}}-\hat{f}^{\mathrm{rfob}})/J_{\mathrm{n}}+\hat{f}^{\mathrm{dis}}/J_{\mathrm{n}},
\label{eq:eq3}
\end{align}
where, $K_{\mathrm{for}}$ represents the force feedback gain. Additionally, the reaction force estimated by the reaction force observer (RFOB)\cite{c21} is denoted as $\hat{f}^{\mathrm{rfob}}$.
\subsubsection{Position/Force control}
Position/force control uses reference values that follow both position and force commands. Therefore, control in two dimensions is achieved by (\ref{eq:eq4}), which adds (\ref{eq:eq2}) and (\ref{eq:eq3}).
\begin{align}
\ddot{x}^{\mathrm{ref}}=&(K_{\mathrm{pos}}+K_{\mathrm{vel}}\frac{d}{dt})(x^{\mathrm{cmd}}-x^{\mathrm{res}})/2 \nonumber\\
&+K_{\mathrm{for}}(f^{\mathrm{cmd}}-\hat{f}^{\mathrm{rfob}})/2J_{\mathrm{n}}+\hat{f}^{\mathrm{dis}}/J_{\mathrm{n}}.
\label{eq:eq4}
\end{align}
The force reference value to the motor is expressed in (\ref{eq:eq5}) by multiplying the moment of inertia by (\ref{eq:eq4}).
\begin{align}
f^{\mathrm{ref}}=&J_{\mathrm{n}}(K_{\mathrm{pos}}+K_{\mathrm{vel}}\frac{d}{dt})(x^{\mathrm{cmd}}-x^{\mathrm{res}})/2 \nonumber\\
&+K_{\mathrm{for}}(f^{\mathrm{cmd}}-\hat{f}^{\mathrm{rfob}})/2+\hat{f}^{\mathrm{dis}}.
\label{eq:eq5}
\end{align}
From the explanation above, it becomes possible to control both the position and force command values. However, suppose the ratio of these command values, representing the motion's impedance, does not match the impedance of the actual contact environment. In that case, controlling the motor with both command values is not feasible. When there is a discrepancy between the two impedances, the controlling impedance, namely the ratio of gains between position control and force control, dictates which command value will be prioritized.

Note that position/force control in this paper refers to regulating either position, force, or both, depending on the control objectives. As used throughout this work, motion denotes the actual movements or actions performed by the robot or motor, such as the trajectories and forces observed during experimental tasks. For example, 'motion data' refers to recorded information on these movements.

\subsection{MRS}

This section discusses the MRS, which reproduces previously acquired motions. In general, in MRS, motion data is initially obtained through bilateral control. Bilateral control is a type of leader-follower system where the follower not only tracks the leader's position but also provides feedback to the leader about the reaction force felt by the follower, enabling bidirectional remote operation\cite{c13, c14}. By operating bidirectionally, the leader can remotely control the follower as if they were directly manipulating it. The position and force data captured in this process include the leader's movements and the adaptability to the environment involving the use of force. Since this paper does not employ bilateral control, detailed explanations of bilateral control are deferred to the references\cite{c13, c14}. This paper used the MRS to control the motor using the position and force measured during a previously executed action.

In MRS, the recorded position $x^{\mathrm{rec}}$ and force $f^{\mathrm{rec}}$ response values are the command values for control. Therefore, the force reference value is indicated by substituting the recorded response value into (\ref{eq:eq5}).
\begin{align}
f^{\mathrm{ref}}=&J_{\mathrm{n}}(K_{\mathrm{pos}}+K_{\mathrm{vel}}\frac{d}{dt})(x^{\mathrm{rec}}-x^{\mathrm{res}})/2 \nonumber\\
&+K_{\mathrm{for}}(f^{\mathrm{rec}}-\hat{f}^{\mathrm{rfob}})/2+\hat{f}^{\mathrm{dis}}.
\label{eq:eq6}
\end{align}
A block diagram of the motor control using MRS is shown in Fig.~\ref{fig:MRS}. In Fig.~\ref{fig:MRS}, unlike a typical MRS, a noise signal is added to the force reference value, and a low-pass filter is applied to the feedback signal. These additions are necessary for the proposed method and were incorporated into MRS to standardize the experimental environment. The reasons for including these components are discussed in Sections II-C and II-D.

However, even with MRS, if the impedance of the motion and that of the environment differ, the motion data is not perfectly reproduced. As mentioned in Section I, this issue often leads to problems such as damage to the object or dropping it. While methods that record multiple sets of motion data and select the motion based on environmental information have been reported\cite{c15}, they still need to be improved due to the finite number of motion data. These methods may only cover some scenarios or struggle with extrapolation, leaving unresolved issues.

\subsection{Physical Property Estimation}
In this paper, motions are generated using environmental impedance. Since physical properties can represent environmental impedance, this section discusses methods for estimating these properties. Physical properties represent the relationship between position and force and are often described by the spring-mass-damper model. The mass $M$ represents inertia, the damper $D$ represents viscosity, and the spring $K$ represents stiffness\cite{c17}. We have already proposed estimating physical properties using four parameters in the spring-mass-damper model with an added load $H$\cite{c18, c19}. The load primarily represents disturbances such as friction or ripple and components of work exerted on the contact object. Hence, the relationship between position $X$ and force $F$ is represented by (\ref{eq:eq7}).
\begin{align}
F(s)=(Ms^{\mathrm{2}}+Ds+K)X(s)+H.
\label{eq:eq7}
\end{align}

Each parameter was sequentially estimated using multiple regression analysis. The details of the algorithm are deferred to the references. The sampling time was 0.1~ms, and the most recent data of 100~ms was used, resulting in a sample size $N$ of 1000.

When estimating physical properties, ensuring the signal's persistence of excitation (PE) is necessary. In other words, if the frequency components of the input signal are sparse, high-order model identification may not be correctly performed. Previous studies have applied a constant disturbance signal, but applying a large disturbance signal at the onset of contact can lead to improper contact due to vibrations\cite{c15}. Conversely, if the disturbance signal is too small, strong contact forces can suppress the disturbance signal, making it impossible to ensure PE. Therefore, an attempt was made to modulate the disturbance signal's amplitude based on the contact disturbance's magnitude. Equations (\ref{eq:eq8}) and (\ref{eq:eq9}) represent the amplitude $f_{\mathrm{amp}}^{\mathrm{noise}}$ and the disturbance signal $f^{\mathrm{noise}}$.
\begin{align}
f_{\mathrm{amp}}^{\mathrm{noise}}=\frac{1}{N}\int_{\mathrm{0}}^{\mathrm{N}} \hat{f}^{\mathrm{dis}}(t) dt,
\label{eq:eq8}
\end{align}
\begin{align}
f^{\mathrm{noise}}=f_{\mathrm{amp}}^{\mathrm{noise}} \{ &\sin(2\pi50t)+\sin(2\pi60t) \nonumber\\
&+\sin(2\pi70t)+\sin(2\pi80t) \nonumber\\
&+\sin(2\pi90t)+\sin(2\pi100t) \}.
\label{eq:eq9}
\end{align}
From the above, the proposed method achieves high-speed physical property estimation by using an RFOB with a control cycle of 0.1~ms. By incorporating a high-frequency noise signal (50 to 100~Hz), it is possible to ensure sufficient excitation for accurate impedance estimation even within a short data window of 0.1~s. The method of imparting disturbance signals and the control system are described in Section II-D.
\begin{figure}[t]
\centering
\includegraphics[width=70mm]{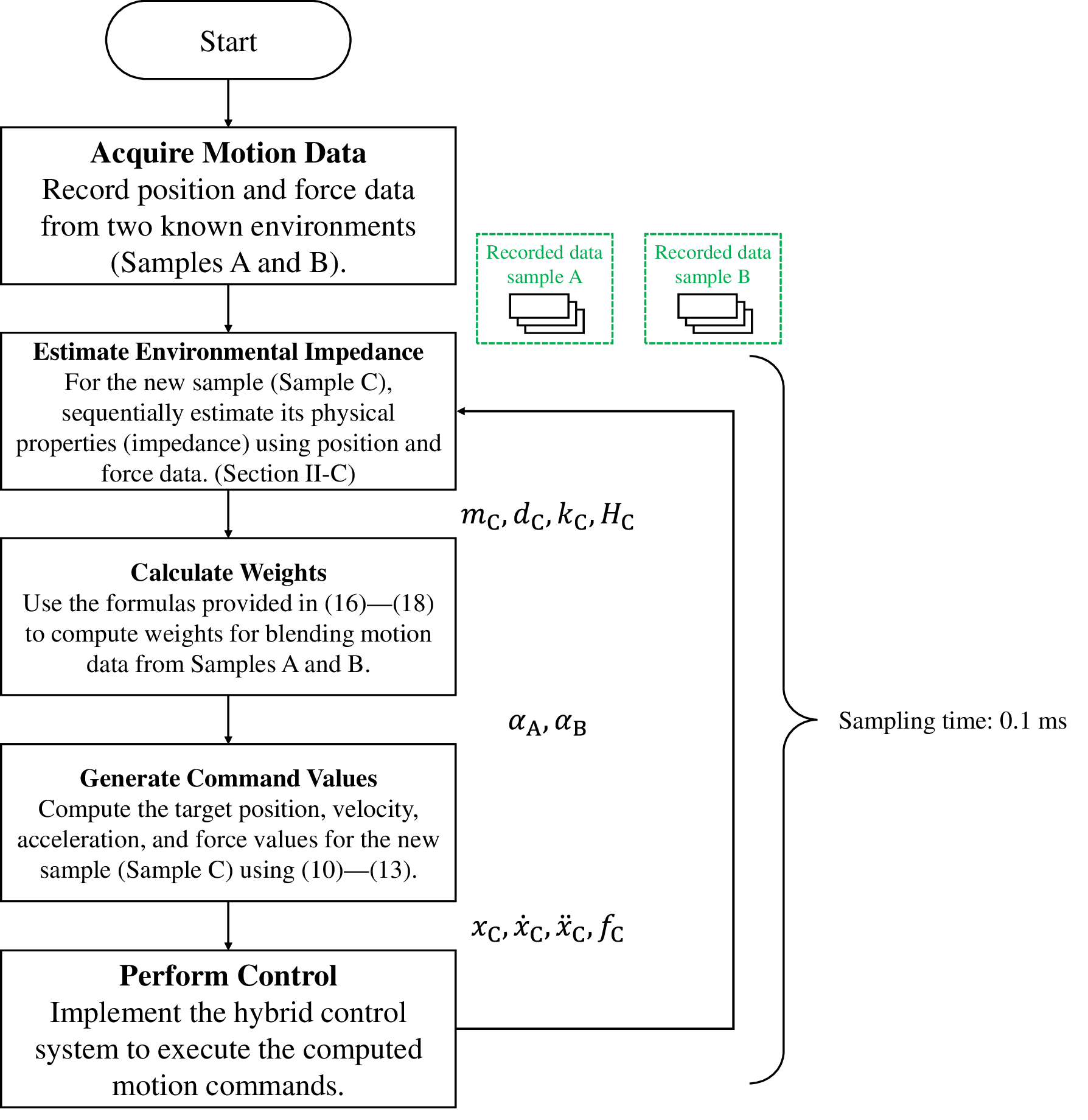}
    \caption{Flowchart illustrating the implementation steps of the proposed method.}
    \label{fig:prop_flow} 
\end{figure}
\subsection{Proposed Method}
This section details the proposed method. The method generates motions by estimating the environmental impedance of a new object based on two sets of motion data previously acquired. In this paper, the samples touched during the acquisition of the two sets of past motion data are referred to as Samples A and B, while the new sample is Sample C. Furthermore, the subscript numbers correspond to the sample numbers. The motion data for Sample C is assumed to be the sum of the motion data for Samples A and B, each multiplied by a weight $\alpha_{\mathrm{A}}$ and $\alpha_{\mathrm{B}}$. Consequently, the target values for position, velocity, acceleration, and force for Sample C are represented by (\ref{eq:eq10})--(\ref{eq:eq13}).
\begin{align}
x_{\mathrm{C}}=\alpha_{\mathrm{A}}x_{\mathrm{A}}+\alpha_{\mathrm{B}}x_{\mathrm{B}}
\label{eq:eq10}
\end{align}
\begin{align}
\dot{x}_{\mathrm{C}}=\alpha_{\mathrm{A}}\dot{x}_{\mathrm{A}}+\alpha_{\mathrm{B}}\dot{x}_{\mathrm{B}} 
\label{eq:eq11}
\end{align}
\begin{align}
\ddot{x}_{\mathrm{C}}=\alpha_{\mathrm{A}}\ddot{x}_{\mathrm{A}}+\alpha_{\mathrm{B}}\ddot{x}_{\mathrm{B}} 
\label{eq:eq12}
\end{align}
\begin{align}
f_{\mathrm{C}}=\alpha_{\mathrm{A}}f_{\mathrm{A}}+\alpha_{\mathrm{B}}f_{\mathrm{B}}.
\label{eq:eq13}
\end{align}
It should be noted that the sum of the weights $\alpha_{\mathrm{A}}$ and $\alpha_{\mathrm{B}}$ is defined to be 1.
\begin{align}
\alpha_{\mathrm{A}}+\alpha_{\mathrm{B}}=1.
\label{eq:eq14}
\end{align}
From (\ref{eq:eq7}), the relationship between the environmental impedance, position, and force in Sample C is represented in the time domain by (\ref{eq:eq15}).
\begin{align}
f_{\mathrm{C}}=m_{\mathrm{C}}\ddot{x}+d_{\mathrm{C}}\dot{x}+k_{\mathrm{C}}x_{\mathrm{C}}+H_{\mathrm{C}}.
\label{eq:eq15}
\end{align}
By substituting (\ref{eq:eq10})--(\ref{eq:eq13}) into (\ref{eq:eq15}) and solving the simultaneous equations with (\ref{eq:eq14}), the values of $\alpha_{\mathrm{A}}$ and $\alpha_{\mathrm{B}}$ can be obtained from (\ref{eq:eq16}) and (\ref{eq:eq17}), and the denominator $den$ of each is given by (\ref{eq:eq18}).
\begin{align}
\alpha_{\mathrm{A}} &= \frac{H_{\mathrm{C}} - f_{\mathrm{B}} + k_{\mathrm{C}} x_{\mathrm{B}} + d_{\mathrm{C}} \dot{x}_{\mathrm{B}} + m_{\mathrm{C}} \ddot{x}_{\mathrm{B}}}{den}
\label{eq:eq16}
\end{align}
\begin{align}
\alpha_{\mathrm{B}} &= \frac{-(H_{\mathrm{C}} - f_{\mathrm{A}} + k_{\mathrm{C}} x_{\mathrm{A}} + d_{\mathrm{C}} \dot{x}_{\mathrm{A}} + m_{\mathrm{C}} \ddot{x}_{\mathrm{A}})}{den}
\label{eq:eq17}
\end{align}
\begin{align}
den =& f_{\mathrm{A}} - f_{\mathrm{B}} + k_{\mathrm{C}} (x_{\mathrm{B}} - x_{\mathrm{A}}) \nonumber\\
&+ d_{\mathrm{C}} (\dot{x}_{\mathrm{B}} - \dot{x}_{\mathrm{A}}) + m_{\mathrm{C}} (\ddot{x}_{\mathrm{B}} - \ddot{x}_{\mathrm{A}}).
\label{eq:eq18}
\end{align}
The block diagram of the proposed method is shown in Fig.~\ref{fig:Prop}. The sequence of control actions for an unknown environment (Sample C) is as follows:
\begin{description}
    \item[\textbf{Pre Step}] \hspace{1em}Acquire motion data using Samples A and B.
    \item[\textbf{Step 1}] Sequentially determine the environmental impedance of Sample C as described in section II-C.
    \item[\textbf{Step 2}] Calculate $\alpha$ from (\ref{eq:eq16})--(\ref{eq:eq18}) based on the motion data of Samples A and B and the environmental impedance of Sample C.
    \item[\textbf{Step 3}] Determine the command values for position and force using the calculated $\alpha_{\mathrm{A}}$ and $\alpha_{\mathrm{B}}$ by (\ref{eq:eq10}) and (\ref{eq:eq13}).
    \item[\textbf{Step 4}] Perform control using the position/force controller described in section II-A.
\end{description}
The flowchart of the proposed method is shown in Fig.~\ref{fig:prop_flow}. Note that Steps 1 to 4 are repeated for each sampling time. Additionally, if there is insufficient data to determine the environmental impedance in Step 1, i.e., if the number of samples is less than $N$, control is performed with $\alpha_{\mathrm{A}}=1$ and $\alpha_{\mathrm{B}}=0$. In addition, if the denominator $den$ is sufficiently small, the value of $\alpha$ becomes indeterminate; therefore, if $den$ is less than 0.001, control is also performed with $\alpha_{\mathrm{A}}=1$ and $\alpha_{\mathrm{B}}=0$. After determining the acceleration reference value in Step 4, as Section II-C outlines, the procedure involves introducing a noise signal to the reference. However, incorporating noise into the position and force readings can distort the impedance estimation. This complication arises because the interplay between input and output (position and force) extends beyond the direct path involving the motor and the gripped object to include the reciprocal path via the controller. To address this problem, our method employs a low-pass filter with a cutoff frequency of 50~Hz for the feedback of position and force values. Filtering out the noise enables the position/force controller to produce an acceleration reference devoid of noise components, thereby facilitating precise impedance estimation for both the motor and the object.

\begin{figure}[t]
\centering
\includegraphics[width=70mm]{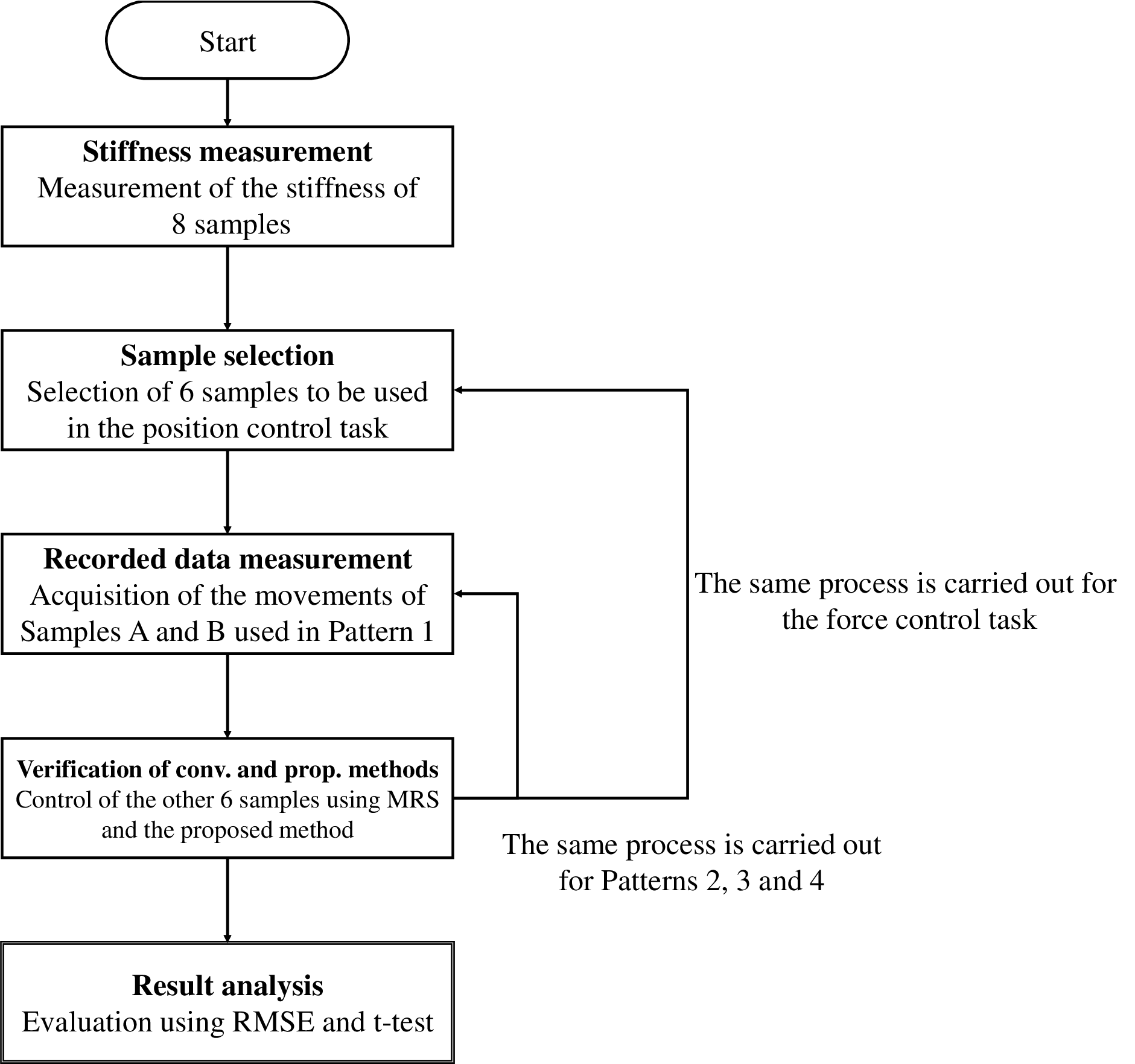}
    \caption{Sequential photographs showing the gripping motion: initial open state, gripping, full grip, and release.}
    \label{fig:exp_flow} 
\end{figure}

\section{Experimental Methods}
The protocol for this experiment is illustrated in Fig.~\ref{fig:exp_flow}. In this experiment, we performed position or force control on two samples beforehand and recorded the data. Using this motion data, we evaluated the error between the original position or force command values and the outcomes when applying MRS to four samples and implementing the proposed method. The two selected samples fall into four patterns: interpolation, extrapolation, and two types of bias. For each pattern, we conducted a t-test to verify the superiority of the proposed method. The details appear in the following subsections.

\subsection{Experimental Setup}

\begin{figure}[t]
\centering
\includegraphics[width=60mm]{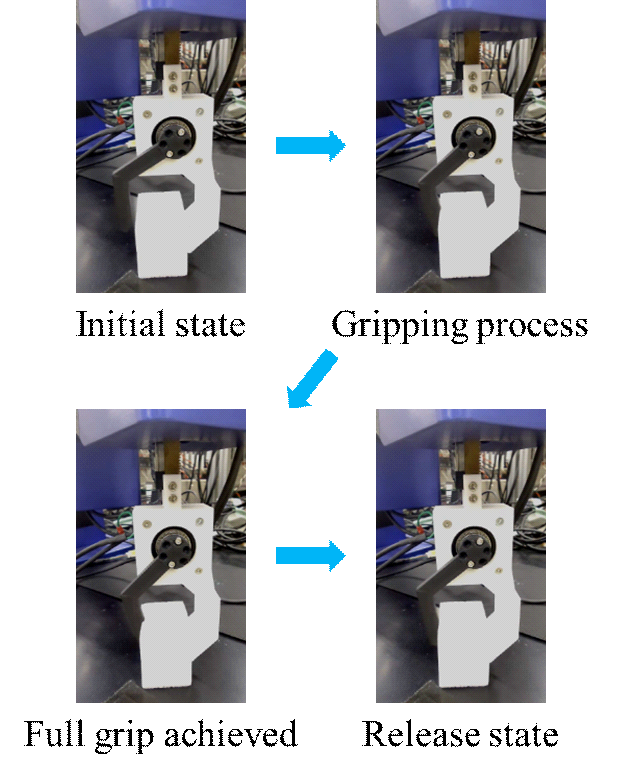}
    \caption{Sequential photographs showing the gripping motion: initial open state, gripping, full grip, and release.}
    \label{fig:Motor} 
\end{figure}

This research developed an experimental apparatus capable of performing gripping actions with a single-degree-of- freedom rotary motor. Fig.~\ref{fig:Motor} depicts the sequential motion of the gripping operation using a series of photographs. In this setup, the white component on the right side of the photograph is fixed to the motor, while the black finger on the left side rotates. The finger's initial position is defined as the moment it makes contact with the sample. This setup ensured continuous contact between the fingers and the object throughout the experiment, preventing sudden changes in environmental impedance during operation. The system could achieve stable impedance estimation and control by maintaining this contact state. The gripping operation was performed based on either position, force, or both command values. The detailed specifications of the motor used are compiled in Table~\ref{tb:motor}. The gripping mechanism adopted a cross-type hand shape \cite{c10}. Using the cross-type hand made it possible to grip soft materials without the fingers interfering with each other.

\begin{table}[t] 
\centering 
\caption{Motor specifications used in the experiment} 
\label{tb:motor} 
\begin{tabular}{|c|c|}
\hline
Type & MDH-4018-6750EG03SH \\
 & (manufactured by Microtech \\
 & Laboratories, Inc.) \\
\hline
Moment of inertia $J_{\mathrm{n}}$ [\rm{$\mathrm{kg} \cdot \mathrm{m}^2$}] & 0.000013589 \\
\hline
Resolusion & 6750 \\
\hline
Gear ratio & 3.0 \\
\hline
Proportional gain $K_{\mathrm{pos}}$ [\rm{$1/s^{\mathrm{2}}$}] & 22500 \\
\hline
Differential gain $K_{\mathrm{vel}}$ [\rm{$1/s$}] & 300 \\
\hline
Force gain $K_{\mathrm{for}}$ [-] & 1.0 \\
\hline
Cutoff frequency of DOB [rad/s] & 800 \\
\hline
\end{tabular}
\end{table}

\begin{figure*}[t]
\centering
\includegraphics[width=130mm]{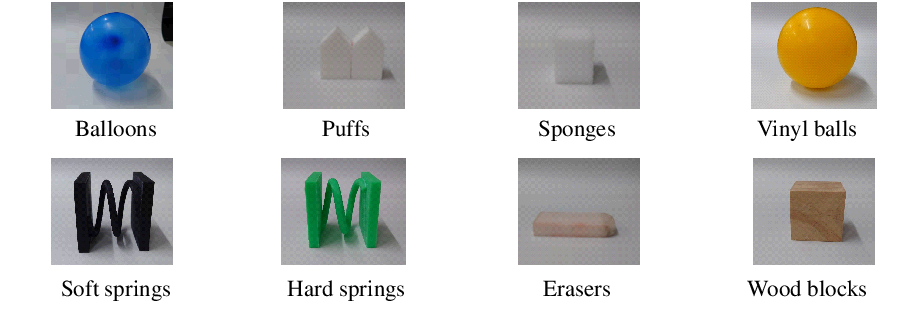}
    \caption{Eight samples used in the experiment}
    \label{fig:Samples} 
\end{figure*}

\begin{table*}[t] 
\centering 
\caption{The sample numbers for each sample are listed in each experiment. If a sample number is not mentioned, it was not used in that particular experiment. When pressed with a force of 0.2 Nm, the stiffness values are also provided.} 
\label{tb:samples} 
\begin{tabular}{|c||c|c|c|c|c|c|c|c|}
\hline
Samples & Balloons & Puffs & Sponges & Vinyl balls & Soft springs & Hard springs & Erasers & Wood blocks \\
\hline \hline
Position control & 1 & 2 & 3 & 4 & 5 & 6 &  &  \\
\hline
Force control & 1 &  & 2 & 3 &  & 4 & 5 & 6 \\
\hline \hline
Stiffness [Nm/rad] & 1.35 & 1.71 & 2.05 & 3.12 & 3.86 & 4.88 & 6.92 & 13.68 \\
\hline
\end{tabular}
\end{table*}
\subsection{Samples}
Eight types of samples with varied impedance characteristics were prepared for the experiments to represent common control challenges in adaptive impedance control and ensure diverse conditions for validation. The eight types of samples are as follows: 
\begin{itemize}
    \item Balloons
    \item Puffs
    \item Sponges
    \item Vinyl balls
    \item Soft springs
    \item Hard springs
    \item Erasers
    \item Wood blocks
\end{itemize} 
The eight samples were selected to cover various impedance characteristics, from low stiffness (e.g., balloons) to high stiffness (e.g., wood blocks), representing common materials that robots may interact with in industrial and practical environments. This selection ensures that the proposed method is tested across diverse operational scenarios, including highly deformable materials and rigid objects. However, using too-hard samples can complicate control in position control experiments. As a result, distinct sets of six samples, chosen from eight, were utilized for the position and force control experiments to address this issue. Details of the eight types of samples are shown in photographs in Fig.~\ref{fig:Samples}, and Table~\ref{tb:samples} lists the samples used in each control experiment along with their corresponding sample numbers.

Additionally, impedance identification tests were conducted for each sample before the experiments. In these tests, the samples were gripped with a force of 0.2 Nm, and the impedance values at that time were measured. Based on the measured impedance, sample numbers were assigned in ascending order of the samples' stiffness.

\subsection{Control Methods}
This paper collected data for both position control and force control, and verification was performed based on the data reproduced in different samples. This section describes the data collected for each type of control.

In the position control experiments, the position command value was set as
\begin{align}
x^{\mathrm{cmd}} = \frac{\pi}{9} (-\cos(2\pi t/10)-\cos(2\pi t/2)+2)\hspace{5pt} [\rm{rad}].
\label{eq:eq19}
\end{align}
Position control was executed such that $x^{\mathrm{cmd}}$ in (\ref{eq:eq2}) became the position command value. The force command value was set as 
\begin{align}
f^{\mathrm{cmd}} = 0.2 (-\cos(2\pi t/10)-\cos(2\pi t/2)+2)\hspace{5pt} [\rm{Nm}],
\label{eq:eq20}
\end{align}
in the force control experiments. Force control was implemented so that $f^{\mathrm{cmd}}$ in (\ref{eq:eq3}) became the force command value. In both cases, the command values were the superposition of two sine waves designed to maintain a gripping state. The recorded motion data is available upon reasonable request. Researchers interested in accessing the data may contact the corresponding author at \textit{t.kitamura@rs.tus.ac.jp}.

\begin{figure*}[htbp]
    \centering
    \begin{subfigure}[b]{0.3\textwidth}
        \centering
        \includegraphics[width=\textwidth]{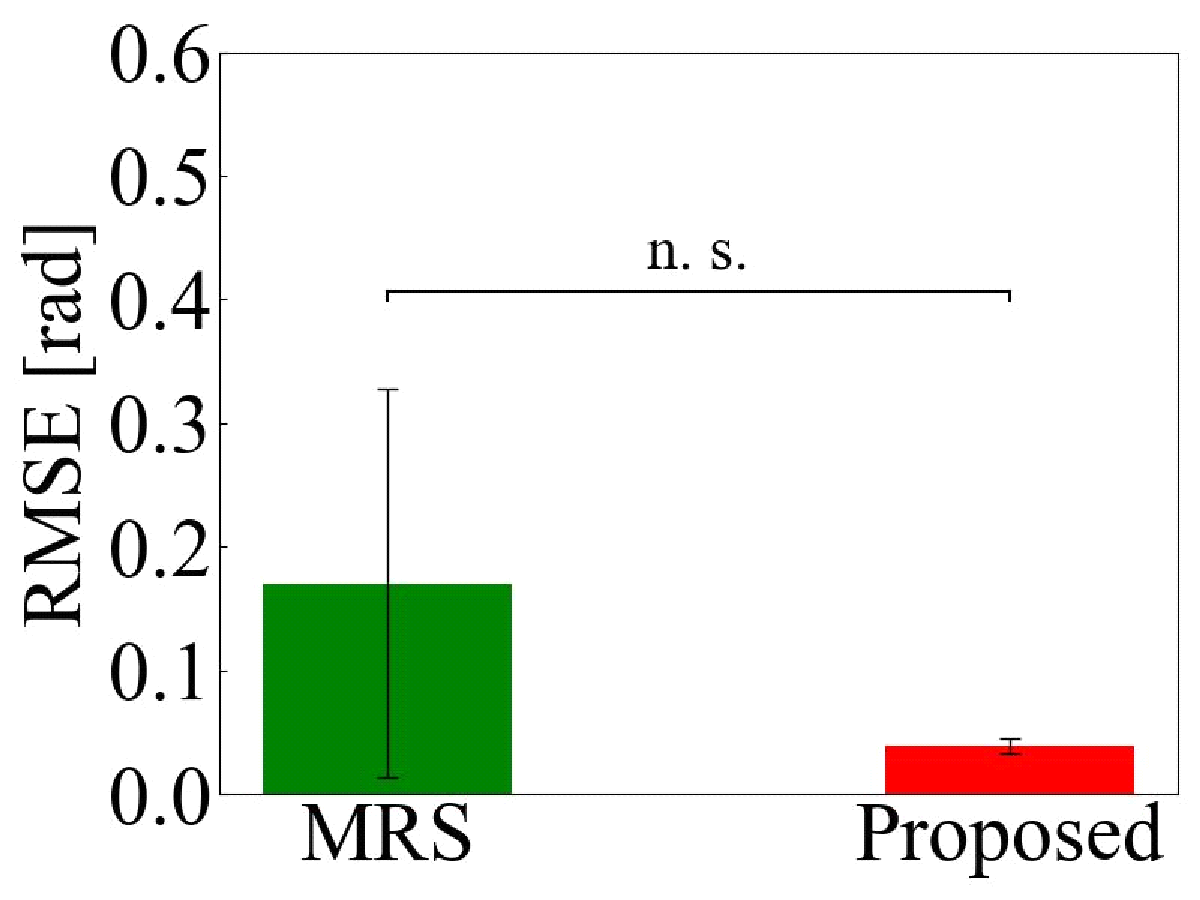}
        \caption{Pattern 1}
        \label{fig:Pos1}
    \end{subfigure}
    \hfill
    \begin{subfigure}[b]{0.3\textwidth}
        \centering
        \includegraphics[width=\textwidth]{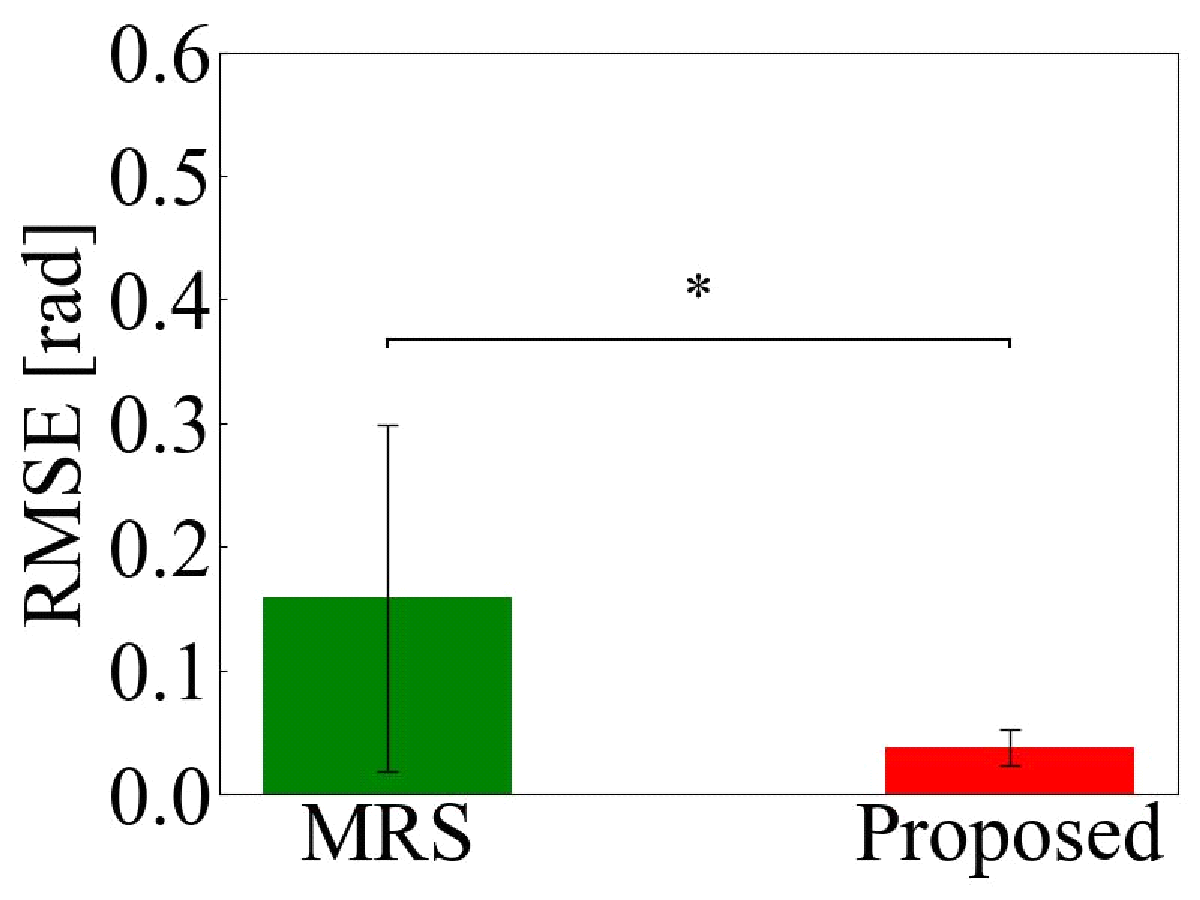}
        \caption{Pattern 2}
        \label{fig:Pos2}
    \end{subfigure}
    \hfill
    \begin{subfigure}[b]{0.3\textwidth}
        \centering
        \includegraphics[width=\textwidth]{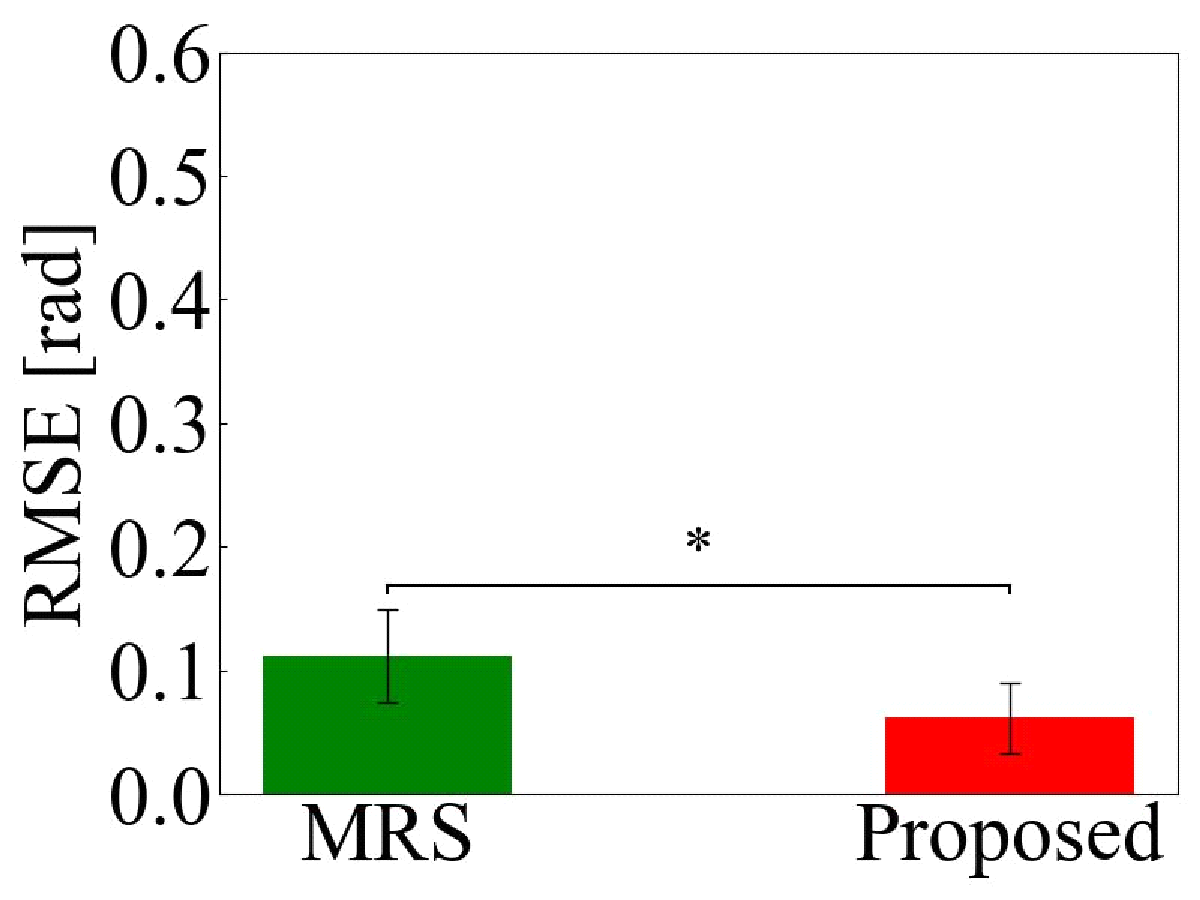}
        \caption{Pattern 3}
        \label{fig:Pos3}
    \end{subfigure}
    
    \vspace*{1cm} 
    \begin{subfigure}[b]{0.3\textwidth}
        \centering
        \includegraphics[width=\textwidth]{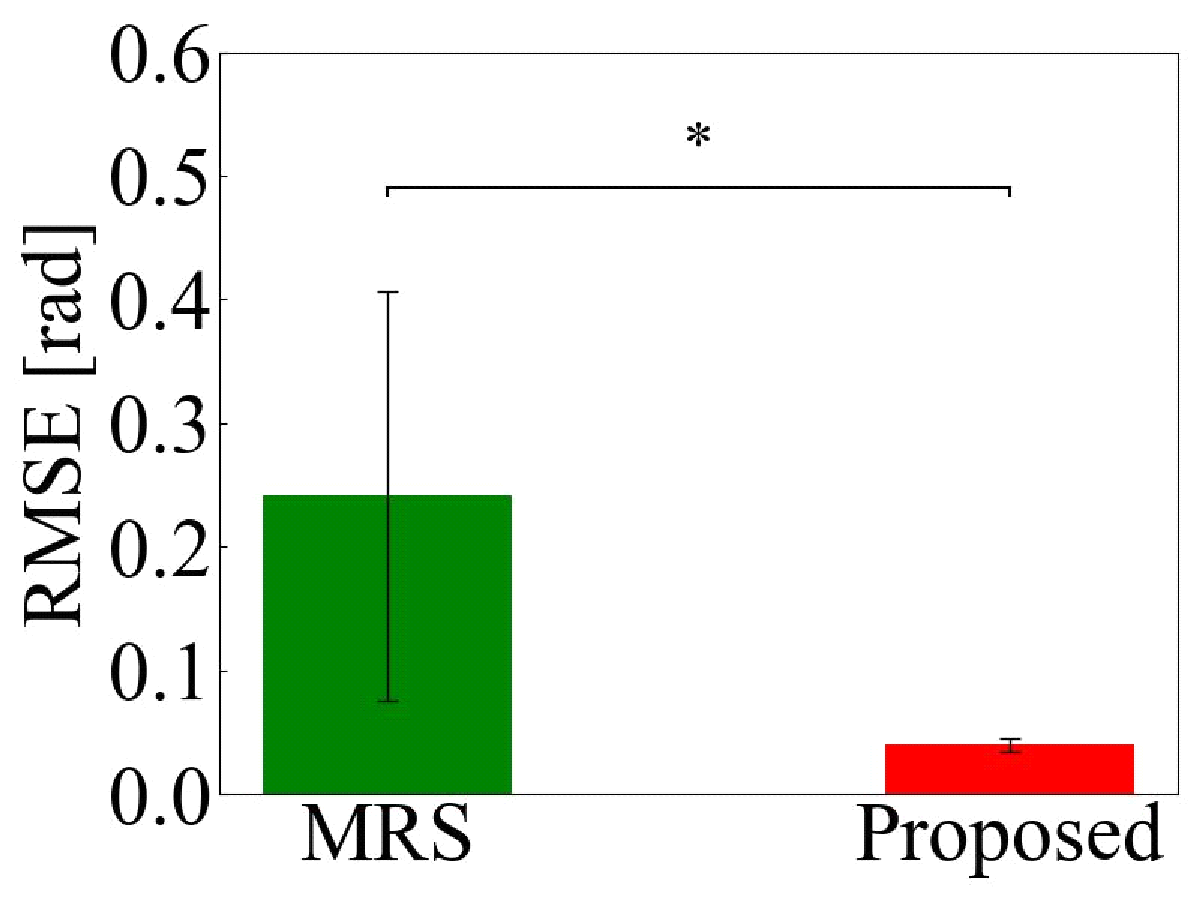}
        \caption{Pattern 4}
        \label{fig:Pos4}
    \end{subfigure}
    \hspace{0.05\textwidth} 
    \begin{subfigure}[b]{0.3\textwidth}
        \centering
        \includegraphics[width=\textwidth]{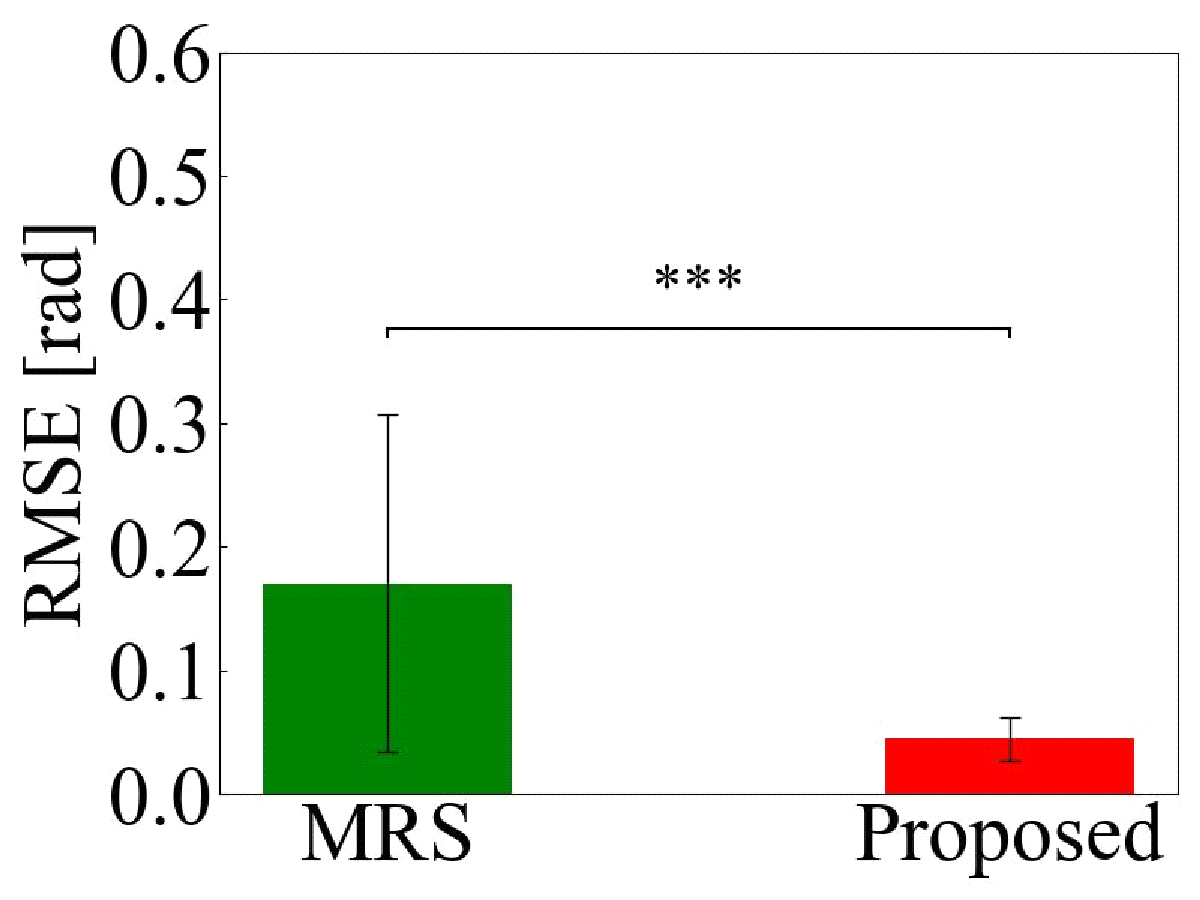}
        \caption{All Pattern}
        \label{fig:Pos5}
    \end{subfigure}
    \caption{RMSE and t-test results based on position control recorded data. (* indicates $p$<0.05, ** indicates $p$<0.01, *** indicates $p$<0.001, and n.s. means $p$$\geq$0.05.)}
    \label{fig:Position}
\end{figure*}

\subsection{Verification Method}
In this paper, we first compared the accuracy of reproducing motion data through position control. Position control was performed on two selected samples (hereafter referred to as Sample A and B), and three types of control were applied to the remaining four samples: using the data from Sample A with MRS, the data from Sample B with MRS, and the data from both Sample A and B with the proposed method. Because of the results of position control on Samples A and B, the original position command values were used for comparison to ensure a fair evaluation. The root mean square error (RMSE) was then calculated based on the original position command values.

The two selected samples were tested in four combinations as follows:
\begin{description}
    \item[\textbf{Pattern 1}] \hspace{15pt}Sample 1 and 6 (the two outermost samples)
    \item[\textbf{Pattern 2}] \hspace{15pt}Sample 3 and 4 (the two innermost samples)
    \item[\textbf{Pattern 3}] \hspace{15pt}Sample 1 and 2 (the two soft samples)
    \item[\textbf{Pattern 4}] \hspace{15pt}Sample 5 and 6 (the two hard samples)
\end{description}
Patterns 1 and 2 correspond to interpolation and extrapolation, respectively. Generally, machine learning methods exhibit unstable behavior for extrapolated actions that have yet to be learned. The proposed method generates actions without learning, and its effectiveness, even for extrapolation, was verified. In Patterns 3 and 4, we tested the capability to generate actions for samples with stiffness levels that differ from those of the samples used to collect the initial motion data.

Secondly, verification was conducted using force control, following the same sample selection as in the position control experiments. In this experiment, six samples, as shown in Table 2, were used. Three types of control were executed, and the force response values were evaluated against the original force command values, namely (20), using RMSE.

\section{Experimental Results}

\subsection{Position Control}
This section discusses the results of action reproduction using position control. The MRS was executed 32 times and the proposed method 16 times, with neither resulting in any instances of control instability. Therefore, the success rate of the gripping action was 100\%. These results confirm that the proposed method retains the environmental adaptability of the MRS.

The RMSE for each pattern, using both the MRS and the proposed method, is depicted in Figs.~\ref{fig:Position}(a)--(d), and the combined RMSE for all patterns is shown in Fig.~\ref{fig:Position}(e). The significance of these errors was assessed using a t-test, revealing a notable difference between the MRS and the proposed method across all patterns, except for Pattern 1. In Pattern 1, the standard deviation of the RMSE for the MRS was larger than that for the other patterns. Pattern 1 was the only pattern with a significant difference in stiffness between the samples. For example, employing Sample 2 as Sample C, the MRS with Sample 1 resulted in smaller errors, whereas the use of Sample 6 resulted in greater errors. This indicates that the standard deviation of the RMSE for the MRS across the four samples increased. Given that the mean and standard deviation of the RMSE for the proposed method in Pattern 1 were not significantly different from those in the other patterns, this suggests that the variability in the MRS results was a key factor in the absence of a significant difference. The results shown in Fig.~\ref{fig:Position}(e) indicate that the average RMSE for the MRS was 0.170~rad, and the average RMSE for the proposed method was 0.0446~rad. Thus, the proposed method reduced the errors by 74\% compared to the MRS.

\subsection{Force Control}
Next, we present the outcomes of reproducing actions using force control data. Similar to the experiments with position control, no instances of motor runaway were observed in the force control trials. Consequently, the success rate of the gripping action reached 100\%, confirming the ability of the proposed method to maintain MRS adaptability to environmental changes, even when using force control data.

\begin{figure*}[htbp]
    \centering
    \begin{subfigure}[b]{0.3\textwidth}
        \centering
        \includegraphics[width=\textwidth]{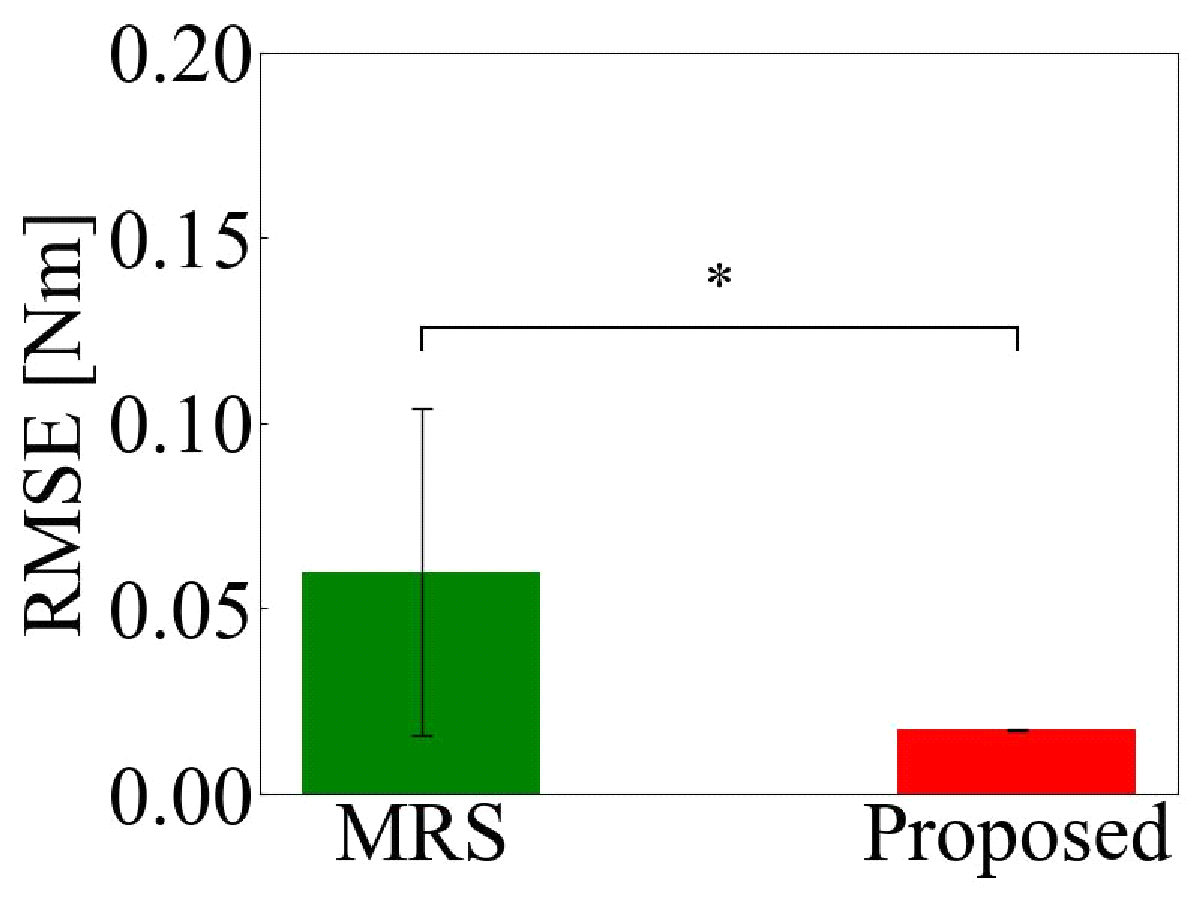}
        \caption{Pattern 1}
        \label{fig:For1}
    \end{subfigure}
    \hfill
    \begin{subfigure}[b]{0.3\textwidth}
        \centering
        \includegraphics[width=\textwidth]{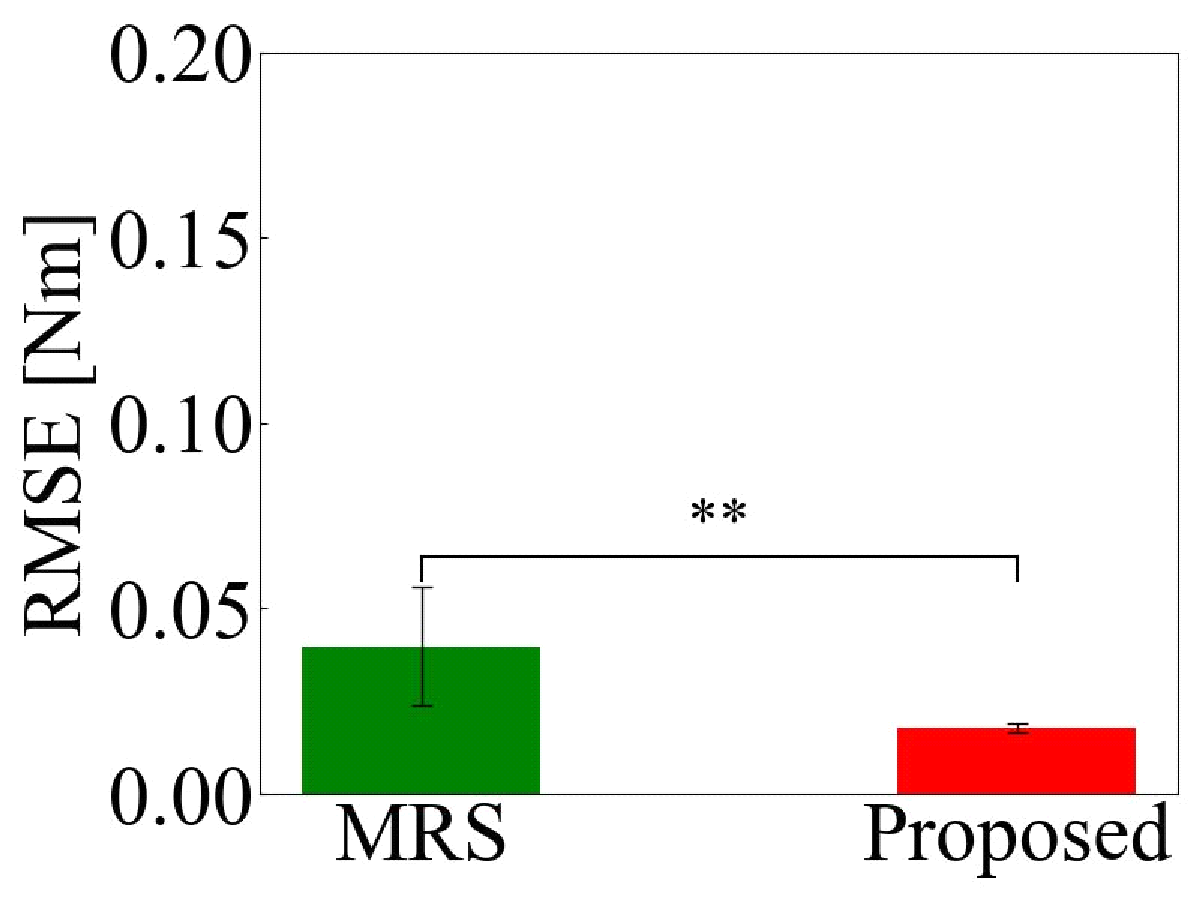}
        \caption{Pattern 2}
        \label{fig:For2}
    \end{subfigure}
    \hfill
    \begin{subfigure}[b]{0.3\textwidth}
        \centering
        \includegraphics[width=\textwidth]{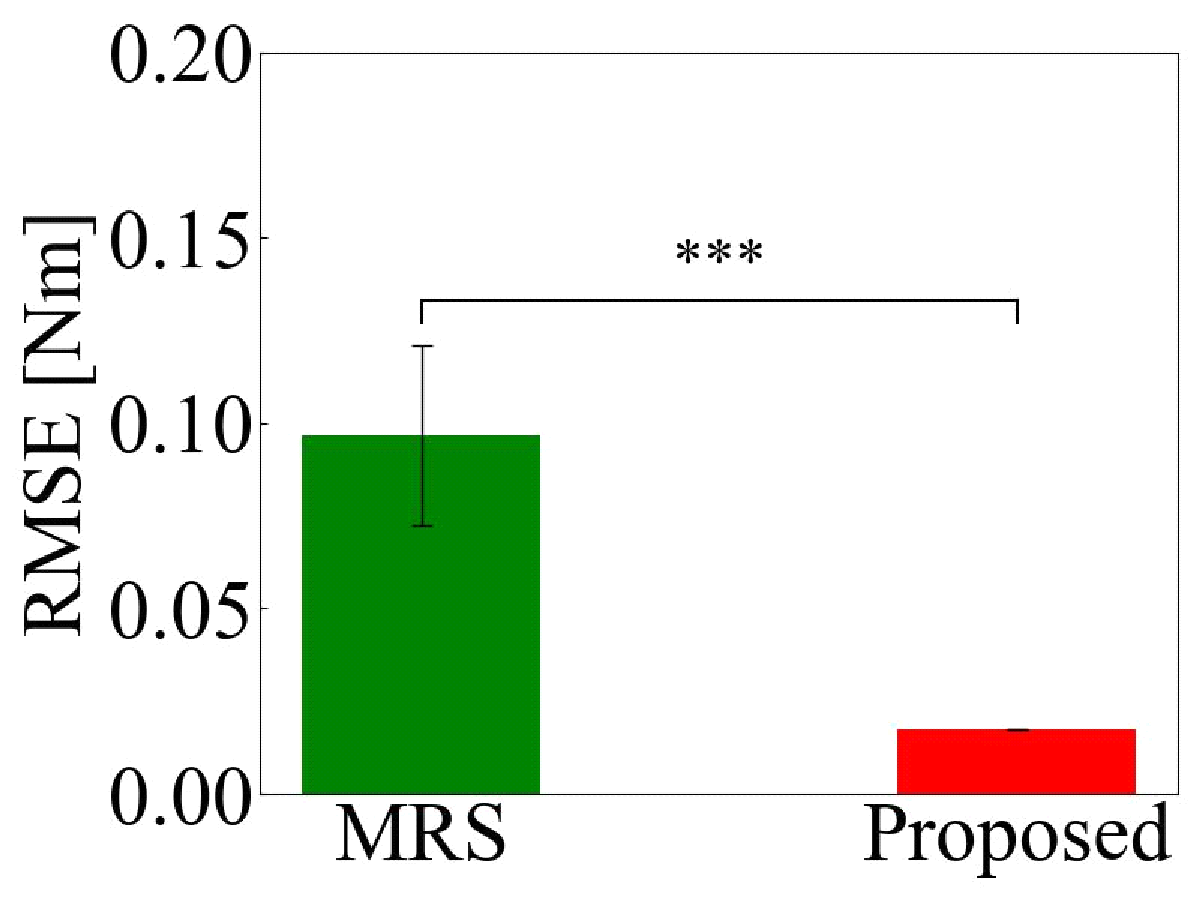}
        \caption{Pattern 3}
        \label{fig:For3}
    \end{subfigure}
    
    \vspace*{1cm} 
    \begin{subfigure}[b]{0.3\textwidth}
        \centering
        \includegraphics[width=\textwidth]{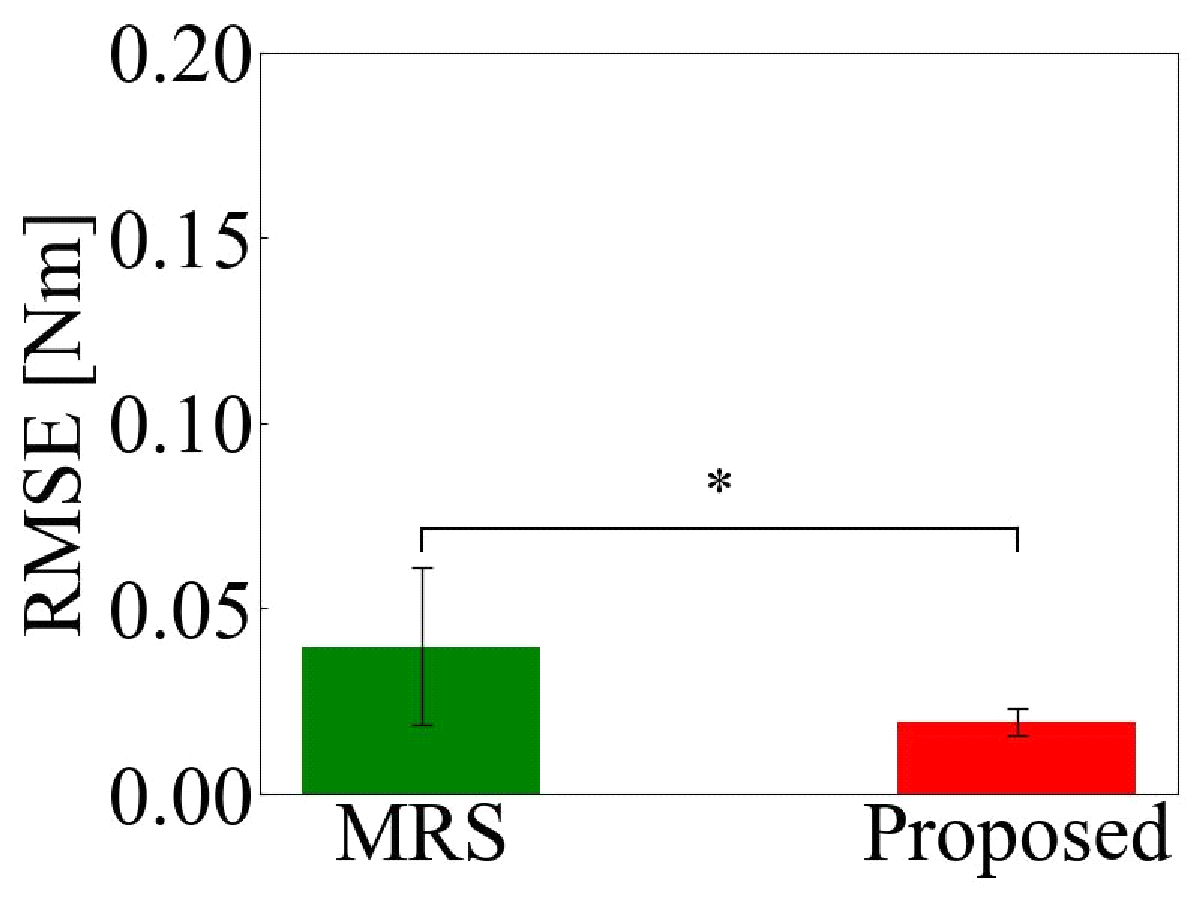}
        \caption{Pattern 4}
        \label{fig:For4}
    \end{subfigure}
    \hspace{0.05\textwidth} 
    \begin{subfigure}[b]{0.3\textwidth}
        \centering
        \includegraphics[width=\textwidth]{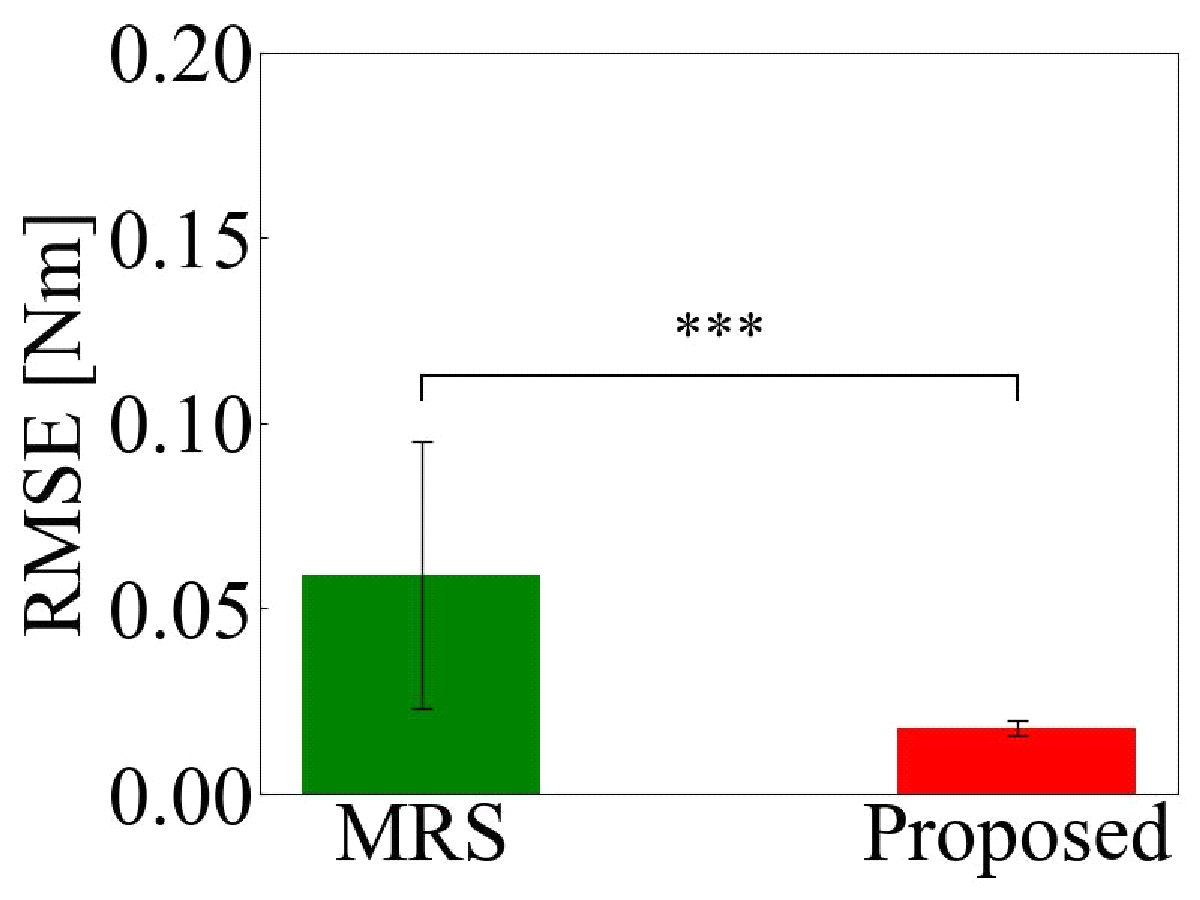}
        \caption{All Pattern}
        \label{fig:For5}
    \end{subfigure}
    \caption{RMSE and t-test results based on force control recorded data. (* indicates $p$<0.05, ** indicates $p$<0.01, *** indicates $p$<0.001, and n.s. means $p$$\geq$0.05.)}
    \label{fig:Force}
\end{figure*}

The RMSE value for each pattern, using both the MRS and the proposed method, is shown in Figs.~\ref{fig:Force}(a)--(d). The aggregate RMSE across all patterns is shown in Fig.~\ref{fig:Force}(e). A significant difference was noted between the performances of the MRS and the proposed method across all patterns, with Pattern 1 showing the most pronounced difference. However, it is important to acknowledge that a direct comparison is challenging due to the use of different samples for the position and force control experiments. The results shown in Fig.~\ref{fig:Force}(e) indicate that the average RMSE for the MRS is 0.0591~Nm, whereas the average RMSE for the proposed method is 0.0180~Nm. Therefore, it was confirmed that the proposed method reduces errors by 70\% compared to the MRS.

\subsection{Discussion}
Through two verifications, it was confirmed that the proposed method significantly improves the reproducibility of actions compared to MRS. MRS attempts to reproduce motion data based on the relationship between the position and force determined by the original target's impedance. In contrast, the proposed method recalculates the position and force according to the current impedance, thereby enhancing reproducibility. The rationale for this approach can be derived from the proposed method, as shown in (\ref{eq:eq10})--(\ref{eq:eq18}). In the ideal case of position control, $x_{\mathrm{A}}$, $x_{\mathrm{B}}$, and $x^{\mathrm{cmd}}$ are equal. The same applies to velocity and acceleration. Consequently, according to (\ref{eq:eq10})--(\ref{eq:eq12}) and (\ref{eq:eq14}), $x_{\mathrm{C}}$ is equal to $x^{\mathrm{cmd}}$. Moreover, by defining $den = f_{\mathrm{A}} - f_{\mathrm{B}}$ and substituting $x_{\mathrm{A}} = x_{\mathrm{B}} = x^{\mathrm{cmd}}$ into (\ref{eq:eq16}) and (\ref{eq:eq17}) and applying the result to (\ref{eq:eq13}) as:
\begin{align}
f_{\mathrm{C}} = m_{\mathrm{C}} \ddot{x}^{\mathrm{cmd}} + d_{\mathrm{C}} \dot{x}^{\mathrm{cmd}} + k_{\mathrm{C}} x^{\mathrm{cmd}} + H_{\mathrm{C}}.
\label{eq:eq21}
\end{align}
Equation (\ref{eq:eq21}) implies that if the motion data for Samples A and B are for position control, then the force command value $f_{\mathrm{C}}$ for Sample C is the force necessary to reach position $x^{\mathrm{cmd}}$, given the environmental impedances $m_{\mathrm{C}}$, $d_{\mathrm{C}}$, $k_{\mathrm{C}}$, and $H_{\mathrm{C}}$. Similarly, for force control, $x_{\mathrm{C}}$ is equal to the position required to exert $f_{\mathrm{C}}$ against the current environmental impedance. Therefore, it was verified that the proposed method generates actions adapted to the current target by matching the impedance of the two motion datasets with the environmental impedance.

Moreover, we confirm the efficacy of the proposed method for samples with an extrapolated impedance. Methods based on learning models typically interpolate training data using approximations. However, beyond the scope of the training data, these methods may generate actions that are misaligned with user intentions in extrapolation scenarios. As described in (\ref{eq:eq21}), the proposed method designs actions influenced by the current environmental impedance, thus facilitating adaptation to environments that are significantly divergent from the initial dataset. The uniform results demonstrated this adaptability for Patterns 2, 3, and 4, where the extrapolation in both position and force control showed no notable variance from the expected outcomes.

This method generates actions based on two sets of preliminary data. In contrast, conventional machine-learning-based motion generation methods require at least several dozen pieces of training data. This difference demonstrates that our method offers significant design, implementation, and practical advantages compared with machine-learning approaches. However, machine-learning methods include learning rough action plans (e.g., recognizing the position of an object and moving forward or to the right) during the learning process. In contrast, our method does not generate the planning part of the action. Therefore, it was necessary to design a simple action planner. An MRS that compensates for variations in the location of objects based on image information has been proposed \cite{c40, c27, c28}. By integrating these, an MRS that adapts to the location and physical properties of an object can be realized. Our method focuses on simplifying the action correction component; however, its usefulness in this field remains significant.

Additionally, a significant advantage of the proposed method is its ability to generate actions without altering existing stable control systems. Traditional methods require adjustments to the control system, such as tuning the gains, making it essential to consider the stability. However, our method assumes a stable control system, thereby eliminating the need for additional stability discussions. The only requirement for applying our method is its ability to measure and control the position and force, making it easily applicable to existing manipulators and mobile units.

\subsection{Limitation}
This study estimated the impedance of the grasped object using multiple regression analysis with a 0.1~s time window, as described in Section~II-C. In the current experimental environment, where the motor continuously performed grasping actions, sudden changes in impedance were not observed. However, in actual use environments, the repetition of contact and noncontact is expected, which means that changes in impedance can occur within the time window. If the impedance changes within the time window, it could lead to a decrease in the estimation accuracy. Therefore, it is necessary to explore new methods for effectively estimating impedance within shorter time windows.

In this study, we demonstrated the efficacy of the proposed method through experiments using a rotary motor with a single degree of freedom (DOF). In addition, the tasks were evaluated using two distinct patterns: position control and force control. Given that this was a foundational study, verification under simplified conditions was necessary. The performance of tasks exhibiting extreme impedance patterns is predicted based on the assumption that similar outcomes can be anticipated for tasks with intermediate impedance values. Future work will focus on verification by using manipulators with multiple degrees of freedom and action data with intermediate impedance levels.

\section{Conclusion}
In this study, we propose a new method to enhance the adaptability of robot actions under diverse environmental conditions by matching the impedance of the actions and the environment. This method obtains the impedance of actions from two sets of prior motion data and sequentially estimates the environmental impedance based on position and force information. Experiments conducted under conditions of differing action impedances, namely, position control (with infinite impedance) and force control (with zero impedance), confirmed that the proposed method can reproduce actions adapted to the current environment.

The most significant advantage of the proposed method is its ability to generate effective actions with less data than machine-learning-based methods. In addition, it can be applied without changes to existing stable control systems. These characteristics suggest that the proposed method has great potential for practical robot applications.

Future research will involve verification by using motion data with intermediate impedance values. For example, this includes motion data in which the impedance is adjusted to grasp firmly when the object is rigid and gently when it is soft. Moreover, the focus will be on developing new estimation methods to effectively address changes in impedance in actions involving both contacts and non-contacts.

\begin{IEEEbiography}[{\includegraphics[width=1in,height=1.25in,clip,keepaspectratio]{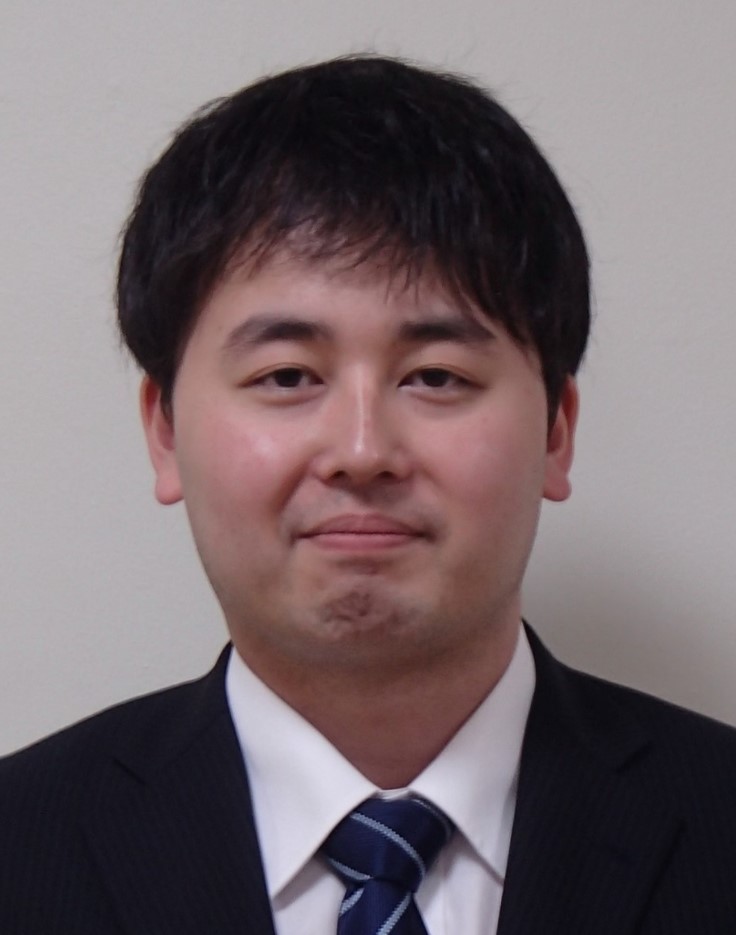}}]{Tomoya Kitamura} received the B.E., M.E., and Ph.D degrees in electrical and electronic systems engineering from Saitama University, Saitama, Japan, in 2015, 2017, and 2021, respectively. From 2021 to 2024, he was a Project Assistant Professor with Keio University, Kawasaki, Japan. Since 2024, he has been an Assistant Professor with Tokyo University of Science, Noda, Japan. His research interests include mechatronics, haptics, and biomedical engineering.
\end{IEEEbiography}

\begin{IEEEbiography}[{\includegraphics[width=1in,height=1.25in,clip,keepaspectratio]{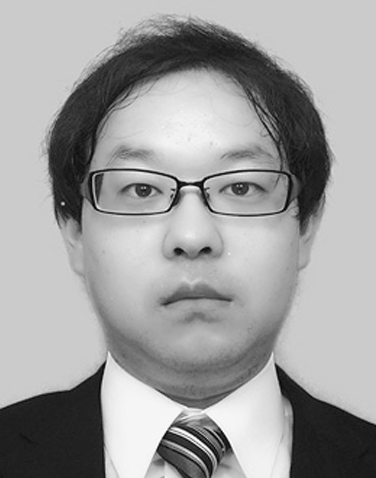}}]{Yuki Saito} received the B.E. degree in system design engineering and the M.E., and Ph.D. degrees in integrated design engineering from Keio University, in 2011, 2013 and 2018, respectively. He has been with Keio University since 2018, where he is currently a Project Assistant Professor. His research interests include motion control, electronics, and haptics.
\end{IEEEbiography}

\begin{IEEEbiography}[{\includegraphics[width=1in,height=1.25in,clip,keepaspectratio]{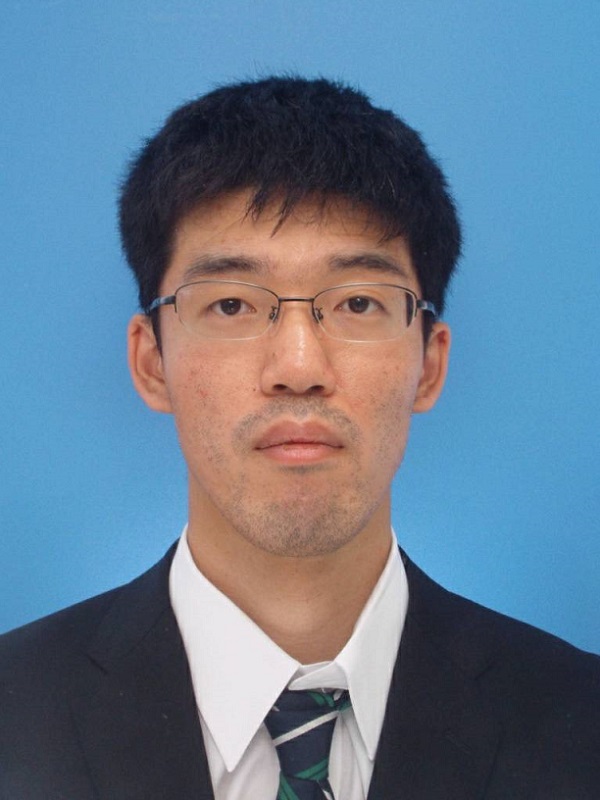}}]{Hiroshi Asai} received the B.E.,M.E.,and Ph.D. degrees in electrical and computer engineering from Yokohama National University, Japan, in 2014, 2016, and 2019 respectively. Since 2019, he has been with Haptics research center in Keio University, Kanagawa, where he is currently an Project Assistant Professor. His research interests include linear permanent magnet machines and motion control.
\end{IEEEbiography}

\begin{IEEEbiography}[{\includegraphics[width=1in,height=1.25in,clip,keepaspectratio]{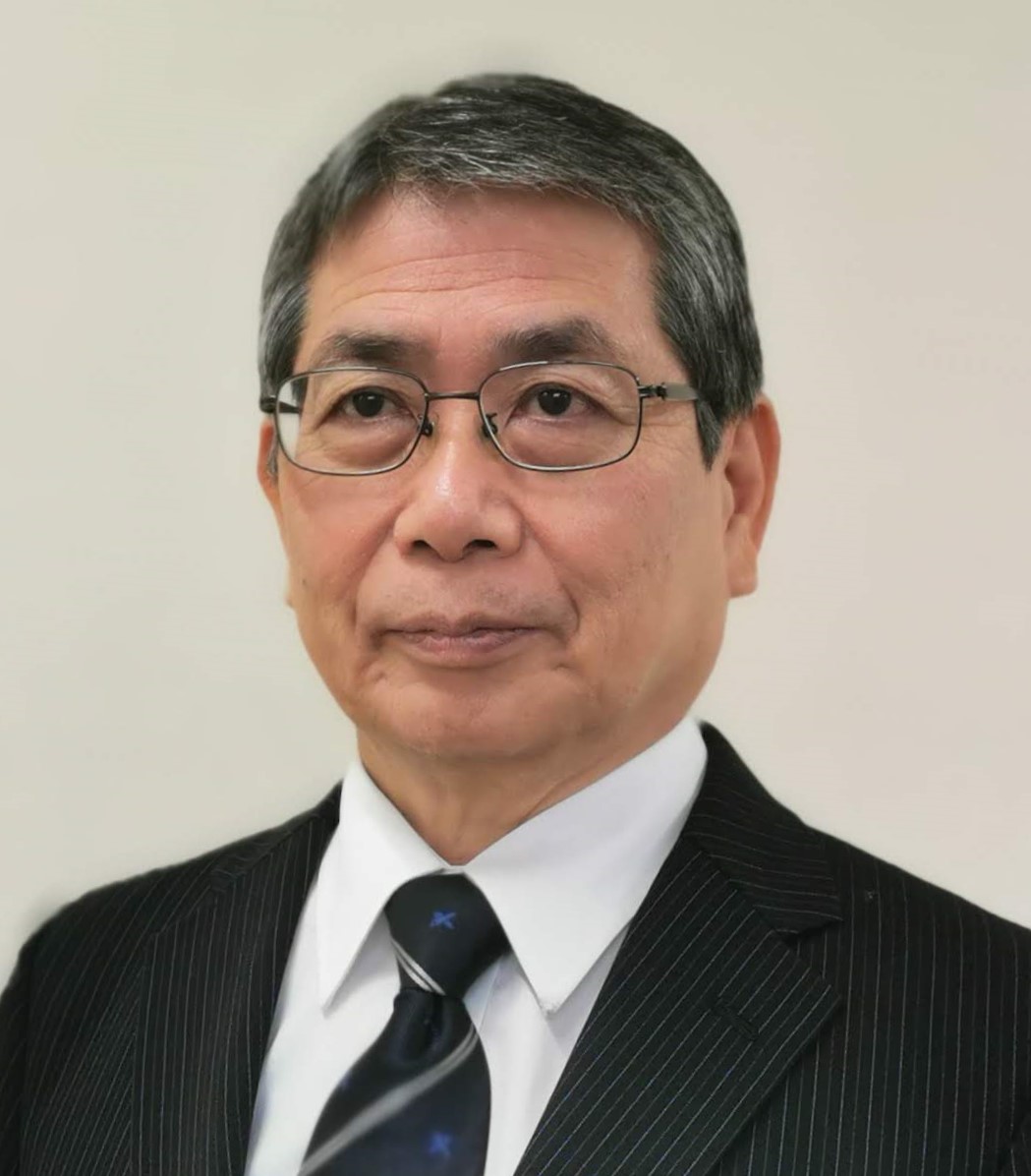}}]{Kouhei Ohnishi} received Ph.D. in electrical engineering from the University of Tokyo in 1980 and has been with Keio University since then. He received IEEJ Outstanding Achievement Award in 2008 and IEEJ Meritorious Contribution Award in 2017. His research field includes motion control, haptics, power electronics, and robotics.
\end{IEEEbiography}

\EOD

\end{document}